# On game psychology: an experiment on the chess board/screen, should you always "do your best", and why the programs with prescribed weaknesses cannot be our good friends?

(**1. Strong opposition during the development stage is necessary for the creation of correct targets; a use of this principle in chess**. **2. Some associated philosophy re human behavior**. **3. The "Chess-Corrida"** )

*Emanuel Gluskin*

Kinneret College on the Sea of Galilee, Braude Academic College, and Electrical Engineering Department of Ben Gurion University of the Negev.  gluskin@ee.bgu.ac.il

**Abstract**:  It is noted that allowing, by means of some specific "unreasonable" moves, a chess program to freely occupy the center of the board, greatly weakens the program's ability to see the serious targets of the game, and its whole level of play.  At an early stage, the program underestimates the ability of the opponent, and by some not justified attack (advance) loses time and helps the other side to reach it in the development.  Weak coordination of Program's figures, caused by quick advance of these figures, is also obvious at this stage.  On a larger scale, the Program is taken out of its library by the unusual start and has difficulties to return to it, often continuing to play indecisively during many of the following moves.  Direct use of these difficulties of the program, and the background psychological nuances, make the play more scientifically attractive and the competition scores gained against the "machine" are also dramatically increased.  The present work is not intended to advance chess learning in the sense of chess art per se, but rather to better (more widely) put this game in the general scope of one's intellectual interests.  This means some general reflections of the problem of keeping/having serious game targets in view of human psychology and education, and the associated modeling, by means of the "unsuccessful" (just as we are) chess programs, of what can happen in the world of human relations and competitions.  It is suggested that program be created with different weaknesses in order to analyze the associated human behavior.  The aspect of competition is also respected, and a specific variation of the game, named "Corrida", based on some variants of the performed experiments is suggested.

## 1. Introduction

### *1.1. General*

The present "intelligence service report" relates to an investigation in the field of the chess game, although the chess *as the art* does not really interest us here, but the psychology of the battle revealed by the analysis of an unexpected weakness of a program that otherwise usually easily defeats me.

Chess is an ancient game:

"*Probably originating in India during or before 7$^{th}$ century, chess spread to Persia, to Arabia, and then to Western Europe". Its name and the term 'checkmate' are sometimes said to derive from the Persian 'shah', "king", and 'shah mat', "the king is dead*". [1]

Let the latter occur only on the chess board, but this game (playing) includes many elements of human psychology which are really interesting: unexpected tactical tricks/combinations, smart strategic decisions, development of long-term plans using the weaknesses of the opponent, gradual enhancement of the position, systematic use of minor advantages, and even knowledge about what the opponent prefers or dislikes ("*I am not playing against wooden pieces*", Emanuel Lasker, Fig.1, right), and some others.

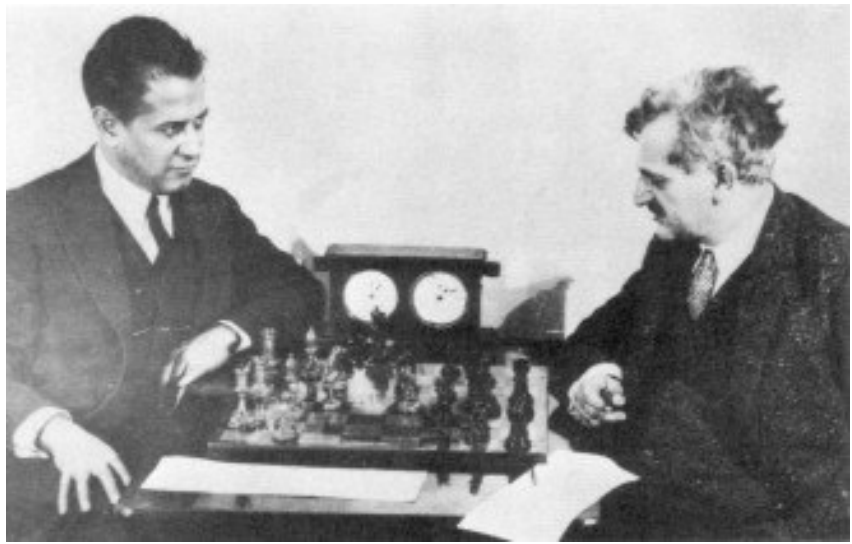

Fig. 1: Hose Raul Capablanca (left) took the chess-crown from Emanuel Lasker, and passed it on to Alexander Alyochin. Each successful champion raised the state of the art of the game to a higher level with the last of which the best modern chess programs, however, successfully compete. However, is the machine-player really as smart as a human one? We argue that this depends on whether or not the human player can, -- unexpectedly, for the machine, i.e. unexpectedly for its Programmer, -- introduce new degrees of freedom in the policy (strategy and/or tactics) of the game. However, the Programmer is, first of all, a Scientist, while the Player is, first of all, a Competitor, and thus it is not a miracle that the machine finally wins. The Player should become a Scientist too, to start to see things more widely, even more philosophically, and the easiest way to cause a Player to become a Scientist is to cause a Scientist (a Mathematician, or a Psychologist, or even a System Theory Specialist) to become, to a degree, a Player.

A keen interest in the high intellectual nature of chess, -- a topic having some relation to our general culture, together with the professional target of automata theory and design, -- led Claude Elwood Shannon in his interesting pioneering works [2-4] to some motivating, even philosophical (in [2] and [4] without any formula), arguments that provided the basis for developing chess programming.

The connection of chess play to human psychology is natural because this very flexible and rich in its possibilities game was invented and developed by humans for themselves. Though this connection is rarely considered, it is the reason for the author's *interest* in the topic and is one of the main focuses in the present *experimental* work. This work is also a logically-critical one, i.e. it criticizes seeing chess play just as a type of competition. Let us, first of all, set our heuristic position in this investigation.

The educational slant of the present work is not so much associated with the victory problem, but much more with a psychological, even philosophical, meaning of the program's observed weaknesses. By analyzing these unexpected weaknesses, we give, in fact, some advice for human education, and finally suggest to the Programmers to creating programs with different kinds of weaknesses, allowing one to model, via the play, the human situations.

The competitive side will be, however, also respected, and based on some specific attempts appearing in our experiment we shall suggest a new dramatic version of chess.

### *1.2. Does the Chess Program really play without "nerves"? Sometimes we shall see the "iron machine" nervous, and sometimes even depressed!*

In [2] and [4] Shannon lists four advantages of the machine over the human player:

*1. Quick counting,*
*2. No mistakes (errors), just some program weaknesses,*
*3. Not lazy,*
*4. No nerves, i.e. no over or under estimations of its position.*

For the last statement, a definition of nervousness seems to be required. The detailed experiment discussed below shows that the programmers can give some nervousness to a machine when programming it to play adventuristically *when it has the impression (in our experiment, induced intentionally) that its opponent is a weak player.* This can be classified as a type of nervousness.

However, this possible nervous play is not the only problem of the program. We also show that if one succeeds, by some very unusual play, in taking the program out of its library, then, as a result of this, it is possible that the program will lose coordination of its figures and will start *and continue, for a long time*, to play much weaker than usual. Isn't this a typical *depression* state? In our experiments such a depression of the program was often observed.

### *1.3. A description of our strategy in simple terms*

It is very difficult to analytically describe the mutual coordination of the actions of the figures. Thus, for instance, considering figures of one color, let us assume that a Knight attacks square *S* of the board, and a Bishop (or another Knight) attacks square *T*, and the Queen attacks both *S* and *T*. Now, let us remove the Queen. That the actions of each of the light figures were coordinated with that of the Queen does not mean that the light figures coordinate with each other, and several moves can be needed for obtaining such coordination. The situation with coordination, is not "transitive", i.e. not as "*if  a = c, and  b = c,  then   a = b*"; forced exchange of one of the figures can destroy the whole coordination.

The following "map" (Fig. 2) demonstrates the start-problems that the Program has. Each '+' means positive influence and each '-' negative one.

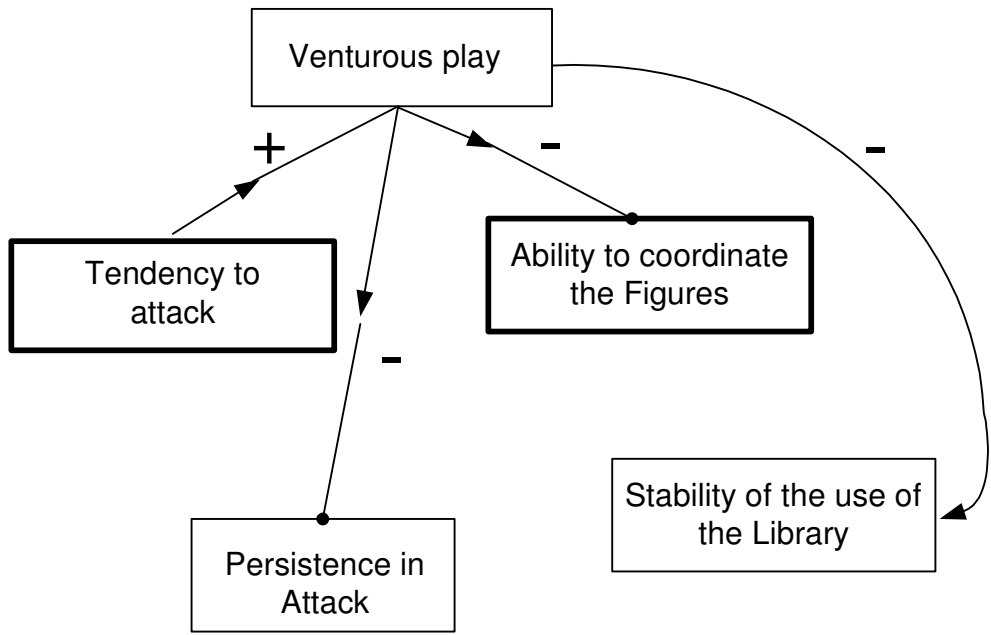

Fig. 2: The map of the problems/features of the Program, observed during the specific start we employ. The directions of the arrows are not arithmetic (algebraic), but logical, i.e. generally non-invertible, and it is not easy to create an analytical theory of even only the start, though in Section 2.13 we do attempt this in very simple terms. These difficulties justify the purely visual ("experimental") description of the start stage, given in this section.

In simple terms, we speak about groups of "weak predators", both for the black and white figures, each of which can be removed and itself be taken, and the *specificity* of the situation under study (i.e. our strategy for White) is as follows:

1. White does not advance figures, letting them to be attacked by Black from distance, and the requirement of closeness of the figures of the opposite groups, for the battle to start, results in a situation in which the advance in space obtained by one side (Black) does not give to this side great advantage, unless Black is lucky to make mate. The latter is, however, not likely because of the confusion in the coordination of Black figures obtained during the *too free* advance of these figures. Thus, White is interested that the real fight should start close to its position.

Though the Program makes its first moves correctly, White is more (very) patient, and Black indeed soon demonstrates poor ability to correctly advance its forces (or create a firm position) in the too luxurious conditions given to it. Since the advanced black figures become poorly coordinated as a whole, and Black starts unjustified attacks that just help White to switch to a quick and easy development, the further play of White does not require high chess skills. (Thus my scores against the machine were drastically improved.)

2. The *initial position* that White reconstructs artificially and unexpectedly for Black, is just very suitable for pursuing the very simple and clear target of starting development when the black figures are close. It sounds paradoxical, but if White had not any immediate trouble, it even can have an advantage at the initial stage of the real battle that thus starts. All this is somewhat similar to the case when one (an analogy for Black) is allowed to freely wave a long sward and attack another man, but if he does not hit him, he soon finds the opponent close to him comfortably operating with a knife. Some other "fight-type" analogies are suggested below in order to stress that the chess psychology is not something isolated understandable only by professionals.

### 1.5. *The role of the coordination of the figures*

It is important to stress that when (as in a usual route of the game) Black is developed with difficulties, it also automatically/necessarily gradually develops *good coordination of its figures*. In terms of the fighting analogies, Black thus takes care to stay on the ground well. When it is developed (advanced) too quickly, then it has poor coordination of its figures, and the period of confusion of Black continues for significant time (measured in the number of the moves). Consider, however, that early unjustified attacks of Black only enhance the coordination problem *that exists here anyway*. For instance, there is no early attacks in Game 6 below, and in several other given games it is also well seen that besides the early attacks, Black has a problem with the coordination of its figures.

*The Program does not see how to use well the possibility of the free (or almost free) movement that White gives to it.*

We turn now to the "laboratory notes and records" of our experiment and to the thoughts regarding its steps and results; a Diary of the Intelligencer. In order to understand the point and feel its romanticism and beauty, the Reader has to use a chessboard, and play out at least some first 20-25 moves in each game considered. Without real watching the game situation it is impossible, for instance, to understand the "corrida"-version of chess, which is one of our final suggestions. Games 1, 4, 5, 6, 7, 8, are, perhaps, most typical, but each of the given games is good evidence of the nervousness and/or depression of the Program in the context of our specific starting tactic-strategy.

Some of the final games with the closed "tracks" of While Knights present the "Corrida" policy in the most clearly, and competitions between humans playing in such style against machine can be suggested.

In general, the games presented in Section 2 give some rich experimental material for a programmer who would wish to find the strategy disadvantages and (mainly) the stability problems for such class of chess Programs as "KChess Elite 4" is.

## 2. From *Alyochin's defense*, to an *Alyochin-type start,* and then to the *"Chess-Corrida"*: the Diary of the Experiment, and the thoughts on line

### *2.1. The observation*

The following observation is not incidental. For a long time I have wanted to check a possible enhancement of the basic idea of Alyochin's defense (**1.** e4 Nf6; **2.** e5 Nd5; **3.** c4 Nb6; **4.** d4 …) in which Black allows White to take the center of the board, and then attacks this center. The point of the defense is that it does not appear to be easy for White to hold the center.
   Undoubtedly, it is very satisfactorily to show to your opponent that his advantage mainly makes him awkward, and I decided to go further with this idea, giving the relevant *initiative* to White (which is generally natural) and letting Black *freely* create its center. This is obtained by White starting with knight(s) (horse(s)) and returning it (them) to the initial place, giving Black some free moves.
   Of course, the chess-program (Black) does not know that this is the policy of White, and starts to play reasonably, i.e. takes the center, not trying to get mate immediately. However when realizing that White plays weakly, Black becomes to be confused in the sense that it cannot choose a correct (serious) plan of the game, and its minor unjustified attacks allow White to quickly advance in his development. Below, we shall analyze this in detail and formulate the things more precisely.
   The most general psychological point is that the whole background psychological potential of the player (as that of Black) can sometimes be developed only while overcoming difficulties starting from the very beginning of the activity.

### *2.2. The experiment*

The "KChess Elite 4" program (free from the Internet for a limited time) plays much better than I do, especially in combinations that the Program finds or initiates much better than I can. Its debut library is also much better than that of mine. When I try to play while "doing my best", then for each case where I win, the program wins some 8-10 games.
   However, after starting my psychological experiment, I was amazed to see that I had a win or a draw much, much more frequently, being almost equal to the program. The three first examples, with only 4 "free moves" in each, follow. Observe in the following three "introductory" games the relatively weak play of Black (the Program) in the period of the "confusion".

The first game

1. Ng1-f3    Ng8-f6
2. Nf3-g1    Nb8-c6
3. Ng1-f3    d7-d5
**4. Nf3-g1**   e7-e5
5. d2-d3     Bf8-c5
6. e2-e3     o-o
7. Ng1-e2    Nf6-g4
8. h2-h3     Qd8-h4
9. g2-g3     Qh4-h5
10. Bf1-g2   Ng4-f6
11. Nb1-c3   Rf8-d8
12. Bc1-d2   a7-a6
13. g3-g4    Qh5-g6
14. Ne2-g3   d5-d4
15. e3xd4    e5xd4
16. Nc3-e4   Bc5-b4
17. Bd2xb4   Nc6xb4
18. Qd1-d2   Nf6-d5
19. a2-a3    Nb4-c6
20. o-o-o    Nc6-e5
21. f2-f4    Ne5-c6
22. f4-f5    Qg6-h6
23. Qd2xh6   g7xh6
24. Ng3-h5   Kg8-h8
25. Rd1-e1   Nd5-e3
26. Rh1-g1   Nc6-e5
27. Nh5-f6   Ra8-a7
28. Ne4-g3   Rd8-d6
29. Ng3-h5   Bc8-d7
30. Bg2-e4   Bd7-a4
31. g4-g5    h6xg5
32. Rg1xg5   Ra7-a8
33. Re1-g1   Ne5-g6
34. f5xg6    f7xg6
35. Be4xg6   h7xg6
36. Rg5xg6   Rd6xf6
37. Rg6xf6   Ne3-f5
38. Rf6xf5   Ra8-g8
39. Rg1xg8+  Kh8xg8
40. Rf5-d5   c7-c5
41. Rd5xc5   Ba4-e8
42. Nh5-f6+  Kg8-f7
43. Nf6xe8   Kf7xe8

44. Rc5-c7    b7-b6
45. c2-c3     Ke8-d8
46. Rc7-h7    d4xc3
47. b2xc3     Kd8-c8
48. Kc1-d2    a6-a5
49. Kd2-e3    Kc8-b8
50. Ke3-d4    Kb8-c8
51. Kd4-d5    Kc8-d8
52. Kd5-e6    Kd8-c8
53. Ke6-d6    a5-a4
54. c3-c4     Kc8-b8
55. Kd6-c6.  Resigns.

The second game

1. Ng1-h3,  Ng8-f6
2. Nh3-g1,  Nb8-c6
3. Ng1-h3,  d7-d6
4. **Nh3-g1**  Bc8-f5
5. Ng1-h3   Nc6-d4
6. d2-d3    Bf5xh3
7. g2xh3    Nf6-d5
8. Bf1-g2   Nd5-b4
9. Nb1-a3   Nb4-c6
10. o-o     e7-e5
11. e2-e3   Nd4-e6
12. c2-c4   Ne6-c5
13. d3-d4   e5xd4
14. e3xd4   Nc5-a6
15. Rf1-e1+ Bf8-e7
16. Bc1-g5  f7-f6
17. Bg5-h4  o-o
18. Na3-c2  Rf8-e8
19. a2-a3   f6-f5
20. Bh4xe7  Re8xe7
21. b2-b4   Re7xe1+
22. Qd1xe1  f5-f4
23. b4-b5   Qd8-g5
24. Qe1-e2  Nc6xd4
25. Nc2xd4  Na6-c5
26. Qe2-g4  Qg5-f6
27. Ra1-d1  Ra8-e8
28. h3-h4   Kg8-h8
29. h4-h5   g7-g6
30. h5-h6   g6-g5

31. Nd4-f5   Re8-f8
32. Rd1-d5   c7-c6
33. b5xc6    b7xc6
34. Rd5xd6   Qf6-a1+
35. Bg2-f1   Nc5-e4
36. Rd6-d7   Qa1-b2
37. Qg4-f3   Qb2-e5
38. Rd7-e7   Ne4-d2
39. Qf3-e2   Qe5xe2
40. Bf1xe2   f4-f3
41. Be2-d3   Nd2-b3
42. Re7xa7   Nb3-c1
43. Bd3-c2   Rf8-d8
44. h2-h3    Nc1-e2+
45. Kg1-h2   Rd8-b8
46. Nf5-d6   Ne2-d4
47. Nd6-e4   Nd4-e6
48. Ne4-f6   Rb8-b7
49. Ra7xb7   Ne6-f8
50. Rb7-b8   c6-c5
51. Rb8xf8*

   The third game

1. Nb1-c3   Nb8-c6
2. Nc3-b1   Nc6-b4
3. Nb1-c3   Ng8-f6
4. **Nc3-b1**   d7-d6
5. Nb1-c3   Bc8-f5
6. d2-d3    e7-e5
7. e2-e4    Bf5-e6
8. Ng1-f3   Bf8-e7
9. g2-g3    o-o
10. Bf1-g2   c7-c5
11. o-o     Qd8-a5
12. Bc1-d2  Qa5-a6
13. Nf3-e1  Nb4xa2
14. f2-f4   e5xf4
15. Bd2xf4  Na2xc3
16. b2xc3   Qa6-b6
17. Ra1-b1  Qb6-c7
18. d3-d4   c5xd4
19. c3xd4   Be6-g4
20. Qd1-d3  Bg4-h5
21. Ne1-f3  Ra8-c8

```
22. Rb1-b2    Bh5-g6
23. Nf3-h4    Qc7-d7
24. Nh4xg6    h7xg6
25. e4-e5     d6xe5
26. Rb2xb7    Rc8-c7
27. Rb7xc7    Qd7xc7
28. Bf4xe5    Qc7-a5
29. Be5xf6    g7xf6
30. c2-c3     Rf8-c8
31. Rf1-c1    Be7-a3
32. Rc1-c2    Rc8-e8
33. h2-h4     Re8-e1+
34. Kg1-h2    Ba3-d6
35. c3-c4     Qa5-h5
36. Kh2-h3    Re1-d1
37. Qd3-e4    Kg8-h7
38. Bg2-f3    Rd1-e1
39. c4-c5     Re1xe4
40. Bf3xh5    Bd6-b8
41. Bh5-f3    Re4xd4
42. c5-c6     Bb8-c7
43. Rc2-b2    Kh7-g7
44. Rb2-b7    Bc7-b6
45. h4-h5     f6-f5
46. h5xg6     Kg7xg6
47. Bf3-e2    Rd4-d6
48. Be2-b5    Rd6-d8
49. g3-g4     f5-f4
50. Bb5-a6    f4-f3
51. Rb7xb6    a7xb6
52. c6-c7     Rd8-f8
53. c7-c8=Q   Rf8xc8
54. Ba6xc8    b6-b5
55. Bc8-a6    Kg6-f6
56. Ba6xb5    f3-f2
57. Kh3-g3    Kf6-g5
58. Bb5-e2    f7-f5
59. g4xf5     Kg5xf5
60. Kg3xf2    1/2-1/2
```

**2.3. *Checking stability of seeing game targets, using the same program (the fourth game)***

The next experiment was as follows. Moving *both* of its knights forward and back, White this time allows Black having not 4, but 6 first free moves. Then, I make several steps (not very few) of my own, and then, not being in any catastrophic situation, let the Program play *for both sides*, assuming that it makes some optimal moves, each time.

In view of the above observations, I was not surprised that White won, because I assumed that Wight's play should be just enhanced by the Program.

In fact, this assumption is not at all simple, and below, based on an example, I have to criticize the play of the program *for any side* when the situation of one side is poorly understood by it. The difficult question of whether or not the ability of the Program to be stable in keeping its game targets can be checked, *using the program itself*, arises.

This is the game:

### The fourth game

1. Nb1-c3   Ng8-f6
2. Nc3-b1   Nb8-c6
3. Nb1-c3   d7-d5
4. Nc3-b1   e7-e5
5. Ng1-f3   e5-e4
6. **Nf3-g1**   Nf6-g4
7. h2-h3    Qd8-h4
8. g2-g3    Qh4-h5
9. e2-e3    Nc6-e5
10. d2-d4   e4xd3
11. c2xd3   Bf8-b4+
12. Nb1-c3  o-o
13. Bf1-e2  Bb4xc3+
14. b2xc3   c7-c5
15. Bc1-a3  Rf8-e8
16. d3-d4   c5xd4
17. c3xd4   Ne5-c4
18. Ba3-c1  Re8-e4
19. Be2-f3  Ng4xf2
20. Ke1xf2  Qh5-f5
21. g3-g4   Qf5-f6
22. Kf2-e2  Re4-e7
23. Bf3xd5  Bc8-e6
24. Bd5xe6  f7xe6
25. Ng1-f3  Re7-f7
26. Rh1-f1  Qf6-h6
27. h3-h4   Rf7-c7
28. e3-e4   Qh6-g6
29. Nf3-g5  Ra8-d8
30. h4-h5   Qg6-e8
31. Ra1-b1  Qe8-c6
32. Ke2-f3  h7-h6
33. Bc1-f4  h6xg5

```
34. Bf4xc7    Rd8-f8+
35. Kf3-g3    Rf8xf1
36. Qd1xf1    Nc4-d2
37. Qf1-c1    Nd2xb1
38. Qc1xc6    b7xc6
39. Bc7-d8    Nb1-a3
40. Bd8-e7    Na3-b5
41. Be7-c5    Nb5-c3
42. Kg3-f3    Kg8-f7
43. Bc5xa7    Nc3xa2
44. Ba7-c5    Na2-c3
45. Bc5-b4    Nc3-b5
46. Kf3-e3    Kf7-f6
47. e4-e5+    Kf6-f7
48. Ke3-d3    g7-g6
49. h5-h6     Nb5-c7
50. Bb4-e7    Nc7-d5
51. Be7xg5    Nd5-c7
52. Kd3-c4    Nc7-b5
53. Kc4-c5    Kf7-g8
54. Bg5-e3    Nb5-c3
55. Kc5xc6    Kg8-f7
56. Be3-f2    g6-g5
57. Bf2-e3    Nc3-e2
58. d4-d5     Ne2-f4
59. Be3xf4    e6xd5
60. e5-e6+    Resigns
```

All the following games, except of the last one in Section 2.11, I again play by my own against Black up to the very end.

## *2.4. Another game with the too early black Queen attack*

In this game, I again let Black to have 6 "free moves". The too early switching the Queen to attack is a typical mistake of the Program in the "overdeveloped" state. But this time, an early Queen attack even led Black, *in a rather late stage* ($27^{th}$ *move*), to lose one of its Knights in order to save the Queen, which points at an *unusual tactical weakness* of the play of the Program that seems to remain for a long time with some problems after the unusual start.

A more general observation is that *the absence of serious targets prevents Black from developing the combination type initiatively-tensioned game in which the Program is much stronger than I am*. I would compare the Program with a human player having a sanguine-type psychological character. Such a person is energetic and patient in any work, even a very difficult one, but only while the proper targets are continuously given to (put before) him.  (The latter is just what I do for the Program when I play normally from the very beginning, honestly "doing my best".)

### The fifth game

1. Ng1-h3    Nb8-c6
2. Nh3-g1    Ng8-h6
3. Nb1-c3    Nh6-g4
4. Nc3-b1    e7-e6
5. Ng1-f3    Bf8-e7
6. **Nf3-g1**    Be7-c5
7. e2-e3     Qd8-h4
8. g2-g3     Qh4-g5
9. Ng1-f3    Qg5-g6
10. Bf1-g2   o-o
11. o-o      Nc6-b4
12. Nb1-a3   Qg6-f5
13. d2-d3    d7-d6
14. Nf3-d4   Bc5xd4
15. e3xd4    Nb4-c6
16. c2-c3    Ng4-f6
17. Na3-c2   Nc6-e7
18. Nc2-e3   Qf5-a5
19. b2-b4    Qa5-a6
20. c3-c4    Qa6-b6
21. Ne3-c2   Ne7-f5
22. c4-c5    Qb6-a6
23. a2-a4    Nf6-d5
24. b4-b5    Qa6-a5
25. Bc1-d2   Nd5-c3
26. Qd1-e1   d6xc5
27. Bd2xc3   Qa5-b6
28. d4-d5    e6xd5
29. Bg2xd5   Rf8-d8
30. Bd5-e4   Nf5-d4
31. Nc2xd4   c5xd4
32. Bc3-b4   Bc8-h3
33. Be4-g2   Rd8-e8
34. Qe1-d2   Bh3xg2
35. Kg1xg2   c7-c5
36. b5xc6    a7-a5
37. Bb4-a3   Qb6xc6+
38. Kg2-g1   Ra8-a7
39. Ra1-c1   Qc6xa4
40. Rf1-e1   Ra7-a8

```
41. Ba3-b2     Re8xe1+
42. Rc1xe1     Ra8-d8
43. Qd2-f4     Qa4-b4
44. Qf4-e5     g7-g6
45. Bb2-a1     b7-b5
46. Kg1-g2     Rd8-d6
47. Re1-e4     f7-f6
48. Qe5-e8+    Kg8-g7
49. Re4-e7+    Kg7-h6
50. Qe8-f8+    Kh6-h5
51. Re7xh7+    Kh5-g4
52. Rh7-h4+    Kg4-f5
53. Qf8-c8+    Kf5-e5
54. Rh4-e4+    Ke5-d5
55. Re4xd4+    Qb4xd4
56. Qc8-b7+    Kd5-e6
57. Ba1xd4     Rd6xd4
58. Qb7xb5     g6-g5
59. Qb5xa5     Rd4xd3
60. Qa5-a6+    Rd3-d6
61. Qa6-c4+    Ke6-f5
62. Kg2-f3     Kf5-g6
63. g3-g4      Rd6-d8
64. Qc4-e4+    Kg6-f7
65. Qe4-f5     Rd8-d6
66. Kf3-g3     Rd6-d1
67. h2-h4      g5xh4+
68. Kg3-f4     Kf7-g7
69. g4-g5      Rd1-d6
70. Kf4-g4     h4-h3
71. Kg4xh3     f6xg5
72. Qf5-e5+    Rd6-f6
73. Qe5xg5+    Kg7-f7
74. f2-f4      Rf6-g6
75. Qg5-e5     Rg6-e6
76. Qe5-d5     Kf7-e7
77. Kh3-g4     Re6-d6
78. Qd5-c5     Ke7-d7
79. Kg4-g5     Rd6-c6
80. Qc5-b5     Kd7-c7
81. f4-f5      Kc7-d6
82. f5-f6.     Resigns
```

## 2.5. *This time White returns to the initial position only at its 8$^{th}$ (eight) move, though in a more nontrivial manner*

It appears possible to come to the initial position even later, -- at the eighth move, though in a less trivial manner, so that the play of Black at this period is somewhat less free (I shall call below such a start as that of "almost free moves"). The following game illustrates that in such a case the Program can play not adventuristically, but very indecisively.

This experiment even suggests reconsidering the opinion that a machine already plays better than a human player does. If I succeed in finding a *successful psychology* against the Program which formally (usually) much stronger than me, -- why cannot Garry Kasparov find something relevant against the machine that once defeated him? Finally, we have a player against a programmer, both humans, and the player has to be not just a strong competitor but also a psychologist, -- against the scientist.

Furthermore, the question of which machine is the strongest also becomes open, while it is not checked whether or not such additional "psychological" degrees of freedom can be used in chess programming.

This is the game:

### The sixth game

1. Ng1-f3     d7-d5
2. Nf3-g5     Nb8-c6
3. Ng5-f3     Ng8-f6
4. Nf3-g1     e7-e6
5. Ng1-f3     Bf8-e7
6. Nf3-h4      o-o
7. Nh4-f3     d5-d4
**8. Nf3-g1**     Nf6-e4
9. d2-d3      Ne4-f6
10. g2-g3     Nc6-b4
11. a2-a3     Qd8-d5
12. Ng1-f3    Nb4-c6
13. Bf1-g2    Nf6-g4
14. o-o       Qd5-b5
15. Nb1-d2    Rf8-d8
16. Nd2-b3    f7-f6
17. e2-e3     d4xe3
18. Bc1xe3    Nc6-e5
19. Nf3xe5    Ng4xe3
20. f2xe3     f6xe5
21. Qd1-f3    Rd8-f8
22. Qf3-e4    Be7-f6
23. a3-a4     Qb5-b6

24. a4-a5      Qb6-d6
25. Nb3-d2    g7-g6
26. Nd2-f3    Qd6-c5
27. c2-c3     Qc5-b5
28. b2-b4     Bc8-d7
29. d3-d4     Bd7-c6
30. Qe4-c2    e5-e4
31. Nf3-e5    Bf6xe5
32. d4xe5     Rf8xf1+
33. Ra1xf1    Qb5xe5
34. c3-c4     a7-a6
35. Qc2-f2    Ra8-d8
36. Qf2-f7+   Kg8-h8
37. Qf7-e7    Rd8-g8
38. Rf1-f7    Rg8-g7
39. Qe7-d8+   Rg7-g8
40. Rf7-f8    Qe5-a1+
41. Bg2-f1    Qa1-g7
42. Rf8xg8+   Qg7xg8
43. Qd8-f6+   Qg8-g7
44. Qf6xe6    Qg7-d7
45. Qe6xd7    Bc6xd7
46. Bf1-g2    Bd7-c6
47. Kg1-f2    Kh8-g7
48. g3-g4     g6-g5
49. Bg2-f1    Kg7-f6
50. b4-b5     a6xb5
51. c4xb5     Bc6-d5
52. a5-a6     b7xa6
53. b5xa6     Kf6-e5
54. a6-a7     h7-h6
55. Bf1-a6    c7-c5
56. Kf2-e2    Bd5-a8
57. Ke2-d2    Ke5-d5
58. Kd2-c3    Kd5-c6
59. Ba6-c8    Kc6-b6
60. Kc3-c4    Kb6xa7
61. Kc4xc5    h6-h5
62. g4xh5     g5-g4
63. Bc8xg4    Ba8-d5
64. Kc5xd5.   Resigns

## 2.6. Again 8 almost free moves, but with an "art experiment" and the resulted strong depression in the play of Black

Let us add an element of art to our strategy. The *symmetric loops* (of a leaf form), the same on each side, right and left, tracked by white Knights before recreating the initial position, make some magic influence on the Program. Black forgets about the necessity to finish developing of its figures, and, at a stage, White even becomes better developed.

Feeling this time very early that my position is already sufficiently strong, I was even not sure in my 13. Nf3xe5, considering instead developing some pressure in the center, but Black soon loses an exchange, becoming inferior in the material. That is, the simple persistent tactic of White *of exchange and simplification* was the best one here too, keeping Black very confused. (See also Section 2.11.)

The whole play of Black is very weak, as if Black continues to think what those symmetric loops by white Knights meant, and remains non-concentrated.

This is the game.

<u>The seventh game</u>

1. Ng1-h3    Ng8-f6
2. Nh3-g5    Nb8-c6
3. Ng5-f3    d7-d5
4. Nb1-c3    d5-d4
5. Nc3-b5    a7-a6
6. Nb5-a3    Bc8-f5
7. Na3-b1    Qd8-d5
**8. Nf3-g1**    Nc6-b4
9. d2-d3    o-o-o
10. a2-a3    Nb4-c6
11. Ng1-f3    Nf6-g4
12. h2-h3    Ng4-e5
13. Nf3xe5    Nc6xe5
14. Bc1-f4    Ne5-g6
15. Bf4-g3    Qd5-b5
16. b2-b3    Ng6-e5
17. Bg3xe5    Qb5xe5
18. Nb1-d2    Qe5-a5
19. e2-e4    Bf5-d7
20. Bf1-e2    Qa5-g5
21. Be2-g4    Kc8-b8
22. Bg4xd7    Qg5xg2
23. Qd1-f3    Qg2xf3
24. Nd2xf3    Rd8xd7
25. Nf3-e5    Kb8-c8
26. Ne5xd7    Kc8xd7
27. f2-f4    f7-f6

```
28. Ke1-e2    e7-e5
29. f4-f5     g7-g6
30. Ra1-f1   Bf8-e7
31. Rh1-g1    g6xf5
32. Rf1xf5   Kd7-e6
33. Rg1-g7   Rh8-c8
34. Rg7xh7   Be7xa3
35. h3-h4    Ba3-c5
36. h4-h5    Bc5-a3
37. h5-h6    Ba3-d6
38. Rh7-g7   Rc8-e8
39. h6-h7    Re8-h8
40. Rf5-h5   Bd6-f8
41. Rg7-g8   Rh8xh7
42. Rh5xh7   Bf8-a3
43. Rh7xc7    b7-b6
44. Rc7-c6+  Ke6-f7
45. Rg8-a8   Ba3-c5
46. Ra8xa6   Kf7-g6
47. Rc6xb6   Bc5xb6
48. Ra6xb6   Kg6-g5
49. b3-b4    Kg5-g6
50. b4-b5    Kg6-g5
51. Rb6-c6.  Resigns
```

Figures 3a,b illustrate the key points.

In Fig. 3a, we have White's initial position "recovered" after 8. Nf3-g1 Nc6-b4, before the forced answer d2-d3. Observe poor coordination of the Black figures; this team does not really know what to do.

In Fig. 3b, we have the position before 25. Nf3-e5 Kb8-c8. That the move Nf3-e5 puts Black in trouble is not the point. The point is that While is already *better developed*, which is obtained by very simple, natural moves, starting from the position in Fig. 3a. Because of the better development, one can objectively (i.e. disregarding the concrete trouble caused by Nf3-e5) prefer the position of While, despite the lack of a pawn. For instance, White can organize a pressure on the Queen-side.

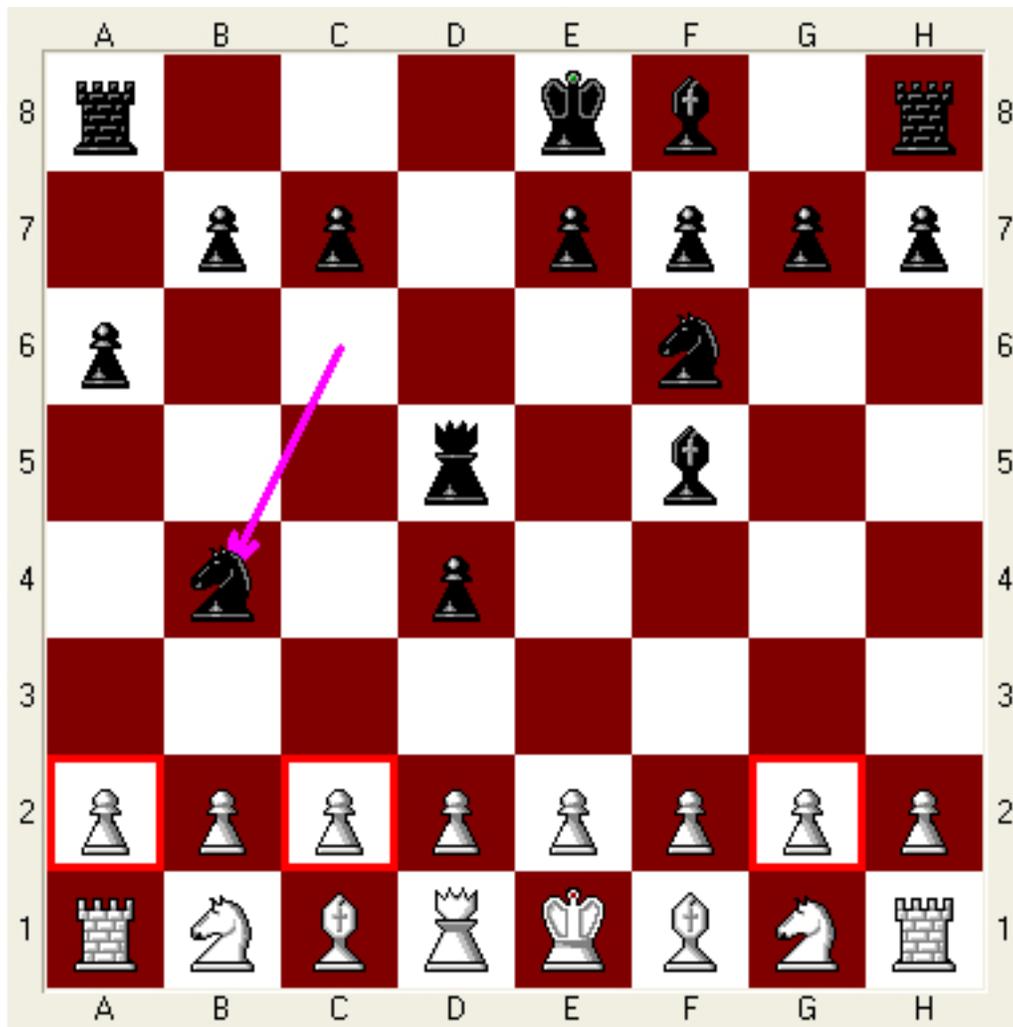

Fig. 3a: The seventh game. The recovered initial White's position, after the leaf-form two-sided loops Ng1-h3-g5-f3-g1 and Nb1-a3-b5-c3-b1. White's move; it will be d2-d3. Coordination of black figures is poor, and though the pawn at d4 is an unpleasant one, they do not form any real dagger or fist.

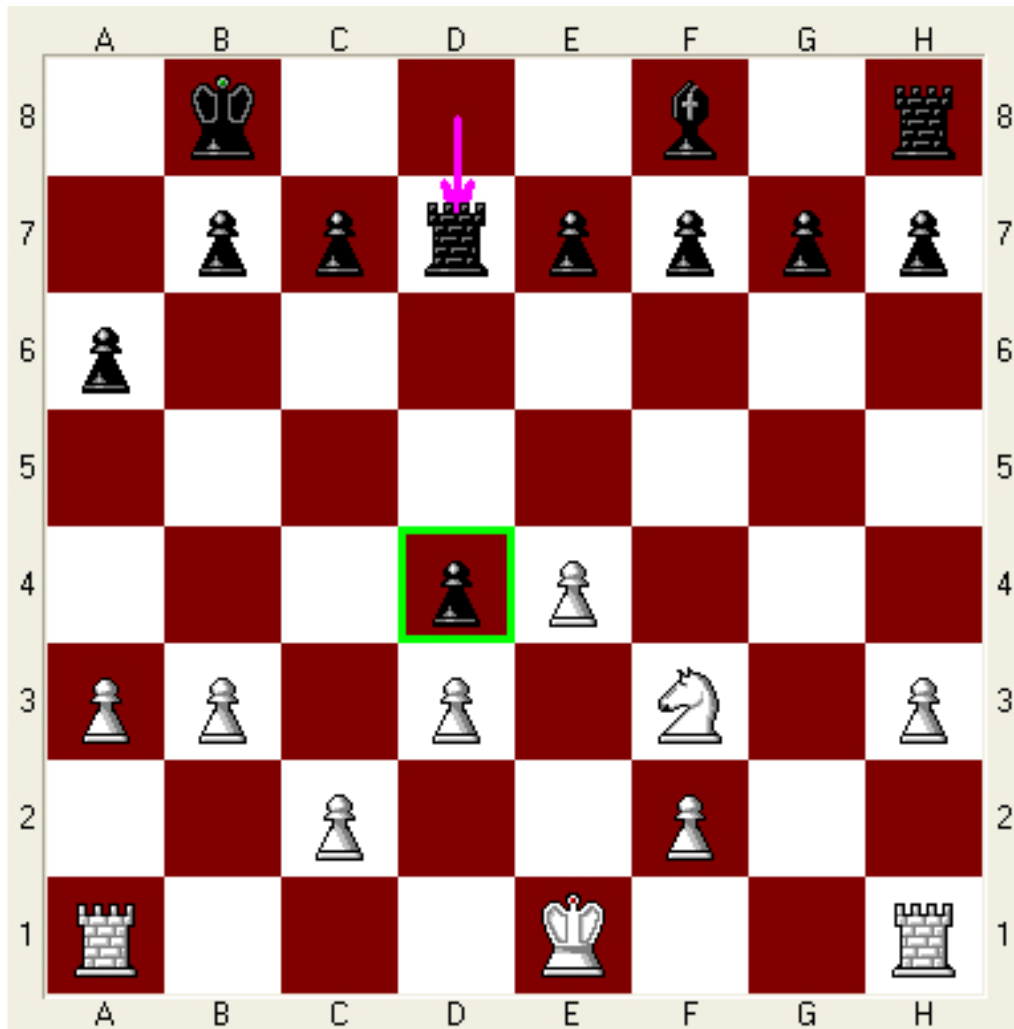

Fig. 3b: The *same* game after 13 moves. Though White lost a pawn, it is better developed. The black pawns' configuration is absolutely unchanged during these 13 moves. The pawn remaining on e7 especially well shows the confusion in the plans of Black during all of the 24 moves passed. If this pawn were to be at e6, Ne5 would not be a great problem. It seems that during these 13 moves Black mainly tried to coordinate its forwarded figures, forgetting about the development of the others. White's simple policy of expelling these forwarded figures and exchanging them made the *programming* target of their coordination *unrealizable* for the Program, and the depression of Black becomes deeper. All this is certainly not just the initial taking the Program out of its debut library; but a very serious decomposition of the power/play of the Program that did not succeed in closing its "hand" (see Fig.3a again) into a fist.

## 2.7. *Some more general observations on line*

1. The seventh and some other games, suggest that the complexity of the program is like the complexity of the set of strings of the piano. That is, one can influence the character of the play of the program in some way by some such art-motives as the symmetric loops of the initial tracks of white knights are. The Programmers, even Shannon himself, hardly thought about such unusual possibilities of creating different levels of confusion of programs. If the Program has its own feeling of art, i.e. some logical impressionability to symmetry and systematicity, this impressionability is a primitive one. The symmetry of the initial Knights' tracks would hardly confuse a human player.

Perhaps, these are too far-reaching terms, not having real chance to survive, but, undoubtedly, this experimentation has interesting research degrees of freedom, some of which should be deeper than it seems at first.

2. I start to notice that the harmful psychological element in the foreground of competition discussed in Section 2.1, which in the usual course of a game, is more weakly exposed in this psychological play against a machine. The psychological "Why?"-s are more interesting than the competition problems. The focus is much more scientific. A new non-harmful application of one's interest to chess is found.

However, let us return to the experiment. Of course, there were games in which Black played well also in the context of the unusual start and I was quickly defeated. Since, however, the Program generally is a much stronger player than I am, none of my failure can be surprising. Let me thus continue only with the cases in which the Program clearly falls out of its main library of serious play and starts to use some simplified (not serious) sub-library.

### 2.8. A game with very early (wrong) decision of the Program that White is a very weak player

The following game is a striking example of Black's switch to such an not serious a sub-library. The move 2.…,Nb4 demonstrates the Program's extremely early decision and the fact that White is very weak. The punishment comes quickly, even for the very careful style of White. Observe the ignorance by Black of the necessity of castling for its King.

<u>The eighth game</u>

1. Ng1-h3 Nb8-c6
2. Nh3-g1 Nc6-b4
3. Ng1-h3 Ng8-f6
4. Nh3-g1 d7-d6
5. Ng1-h3 Bc8-f5
6. Nb1-a3 Nf6-e4
7. Nh3-g1 e7-e5
8. Ng1-f3 Bf5-e6
9. e2-e3 Nb4xa2
10. Bf1-e2 Na2xc1
11. Ra1xc1 Be6-g4
12. o-o f7-f5
13. h2-h3 Bg4-h5

    14. d2-d3 Ne4-g5
    15. Nf3xg5 Bh5xe2
    16. Qd1xe2 Qd8xg5
    17. f2-f4 Qg5-g6
    18. f4xe5 d6xe5
    19. Qe2-f3 Bf8xa3
    20. b2xa3 Qg6-g5
    21. Qf3xf5 Qg5xe3+
    22. Kg1-h1 Ke8-d8
    23. Rc1-e1 Qe3-g3
    24. Re1xe5 c7-c6
    25. Qf5-e6 Qg3-g6
    26. Qe6-e7+ Kd8-c8
    27. Rf1-f7 Qg6xf7
    28. Qe7xf7 b7-b6
    29. Re5-e7 Rh8-d8
    30. Re7-c7 + Kc8-b8
    31. Rc7-b7+ Kb8-c8
    32. Qf7-c7* 1-0.

### *2.9. Back to the initial "art-tracks" by white knights, now performed in parallel; Black plays better but its advantage in the development disappears as quickly as usual*

This was a difficult game, showing that 8 "almost free" moves are close to the boundary of the unusual "generous" strategy that can be chosen by White.

    The ninth game

    1. Ng1-f3    Ng8-f6
    2. Nb1-c3    Nb8-c6
    3. Nf3-g5    e7-e5
    4. Nc3-b5    h7-h6
    5. Ng5-h3    a7-a6
    6. Nb5-a3    d7-d5
    7. Nh3-g1    Nf6-e4
    **8. Na3-b1**    Bf8-c5
    9. e2-e3    Qd8-h4
    10. g2-g3    Qh4-d8
    11. Bf1-g2    o-o
    12. d2-d3    Ne4-f6
    13. Nb1-d2    Bc8-g4
    14. f2-f3    Bg4-e6
    15. Nd2-b3    Nf6-d7

16. Nb3xc5   Nd7xc5
17. Ng1-e2   Nc6-b4
18. o-o      Be6-f5
19. a2-a3    Nb4-c6
20. e3-e4    d5xe4
21. d3xe4    Bf5-e6
22. Bc1-e3   Qd8-e7
23. Ne2-c3   Ra8-d8
24. Qd1-e2   Nc6-d4
25. Be3xd4   e5xd4
26. Nc3-d1   d4-d3
27. c2xd3    Nc5xd3
28. Nd1-f2   Qe7-c5
29. Kg1-h1   Nd3-e5
30. Ra1-c1   Ne5-c4
31. b2-b3    Qc5-e3
32. Qe2xe3   Nc4xe3
33. Rf1-e1   Ne3xg2
34. Kh1xg2   Rd8-d7
35. b3-b4    Rf8-e8
36. h2-h4    Kg8-f8
37. g3-g4    Re8-d8
38. f3-f4    Be6-b3
39. e4-e5    Bb3-e6
40. f4-f5    Be6-d5+
41. Kg2-g3   Bd5-c6
42. g4-g5    h6xg5
43. h4xg5    Rd7-d5
44. Kg3-g4   Rd5-d2
45. Rc1-d1   Rd2xd1
46. Re1xd1   Rd8xd1
47. Nf2xd1   g7-g6
48. f5xg6    f7xg6
49. Nd1-c3   Kf8-e7
50. Nc3-d1   Ke7-e6
51. Kg4-f4   Ke6-d5
52. Nd1-e3+  Kd5-e6
53. Ne3-c2   Ke6-d5
54. Nc2-e3+  Kd5-e6
55. Ne3-c2   Ke6-d5
56. Nc2-e1   Kd5-c4
57. Ne1-f3   Kc4-b3
58. e5-e6    Kb3xa3
59. Nf3-e5   Bc6-b5
60. Ne5xg6   Ka3xb4
61. Ng6-e5   Bb5-a4

   62. g5-g6  c7-c5
   63. g6-g7  Resigns

***2.10. White returns to the initial position only at the 10th move, the position soon appearing is closed and simple. In general, Black plays well, and due to its very clear defense targets, White plays satisfactorily. The game becomes "usual", but having already many figures exchanged, White succeeds to achieve a draw. Ten "almost free" moves are considered to be the maximum for any reasonable experiment***

In the following tenth game we "jump over" the period of the uncertainty, i.e. over all the positions that for the Program are without any "best move". For the 10 "almost free" moves given to Black, the period of its uncertainty and depression already become irrelevant. As a rule, Black has the time to be normally developed and to organize a crucial attack.

 In terms of the time functions ("in other words"), we can say that while in the previous games, there is a "singularity" in development of the game at the moment when White started to play normally, in the game with the maximal number of strange moves, the development of the game becomes "smooth", almost as in a usual game (no real "shock" for Black). This is a direct evidence that 10 is a maximal number of the strange moves.

 Though also in the present game there is no very serious "cavalry" attack of Black, helping White as usual, on the whole, the advance of the black figures, occurring during these 10 moves, is systematic, very massive, and we come to a sufficiently closed and "well-defined" position in which Black successfully tries to increase the pressure, while White has the simple usual defense targets, which helps it to play sufficiently well in order to achieve a difficult draw. As usual, in order to simplify the situation, White tends to exchange the figures, and, fortunately, the position becomes open too late for Black to show its combinational force.

    The tenth game

   1. Nb1-c3  Ng8-f6
   2. Nc3-b5  Nb8-c6
   3. Ng1-f3  a7-a6
   4. Nb5-a3  d7-d5
   5. Na3-b1  e7-e6
   6. Nf3-h4  Bf8-d6
   7. Nh4-f3  o-o
   8. Nf3-g1  Nc6-b4
   9. Nb1-c3  d5-d4
   **10. Nc3-b1**  Nf6-e4
   11. d2-d3  Ne4-c5
   12. Ng1-f3  e6-e5
   13. g2-g3  Bc8-g4

```
14. Bf1-g2    f7-f5
15. o-o       Bg4xf3
16. e2xf3     Qd8-d7
17. a2-a3     Nb4-d5
18. Nb1-d2    Qd7-f7
19. Nd2-b3    Nc5xb3
20. c2xb3     f5-f4
21. Rf1-e1    f4xg3
22. h2xg3     Qf7-f5
23. Qd1-d2    c7-c5
24. Re1-e4    Rf8-f7
25. Qd2-g5    Ra8-f8
26. Qg5xf5    Rf7xf5
27. Bc1-d2    b7-b6
28. Ra1-c1    Bd6-c7
29. Rc1-e1    b6-b5
30. Kg1-f1    Rf5-h5
31. g3-g4     Rh5-h4
32. Bd2-g5    Rh4-h2
33. Kf1-g1    Rh2xg2+
34. Kg1xg2    h7-h6
35. Bg5-d2    Nd5-f6
36. Re4xe5    Bc7xe5
37. Re1xe5    Nf6-d7
38. Re5-d5    Nd7-f6
39. Rd5xc5    Rf8-e8
40. Kg2-f1    Re8-f8
41. Rc5-c6    Nf6-d7
42. Rc6xa6    Nd7-c5
43. Ra6-b6    Nc5xb3
44. Bd2-b4    Rf8xf3
45. Rb6xb5    Rf3xd3
46. Kf1-e2    Nb3-c1+
47. Ke2-f1    Rd3-d1+
48. Kf1-g2    Kg8-h7
49. a3-a4     Nc1-d3
50. a4-a5     Nd3xb2
51. a5-a6     Rd1-a1
52. Bb4-a5    d4-d3
53. a6-a7     d3-d2
54. a7-a8=Q   Ra1-g1+
55. Kg2xg1    d2-d1=Q+
56. Kg1-g2    Qd1xg4+
57. Kg2-h1    Qg4-c4
58. Qa8-d5    Qc4-f1+
59. Kh1-h2    Qf1xf2+
```

```
60. Qd5-g2    Qf2-h4+
61. Qg2-h3    Qh4-e7
62. Qh3-f5+   g7-g6
63. Qf5-e5    Qe7-h4+
64. Kh2-g1    Qh4-g4+
65. Kg1-h1    Qg4-h3+
66. Qe5-h2    Qh3-f1+
67. Qh2-g1    Qf1xb5
68. Qg1-a7+   Kh7-g8
69. Qa7-a8+   Kg8-f7
70. Qa8-f3+   Kf7-e6
71. Qf3-e4+   Ke6-d7
72. Qe4-d4+   Kd7-c8
73. Qd4-c3+   Nb2-c4
74. Ba5-b4    g6-g5
75. Qc3-h3+   Qb5-d7
76. Qh3xh6    Qd7-b7+
77. Kh1-h2    Qb7xb4
78. Qh6xg5    Qb4-d2+
79. Qg5xd2    Nc4xd2
80. 1/2-1/2.
```

## 2.11. Another such game; the helpful role of the tracks of white Knights suggests a new ("corrida") variant of chess

The next *game* also employing 10 "almost free moves" is somewhat different, because the long tracks of white knights "psychologically" caused Black to organize a sufficiently serious attack, and I was again lucky with a difficult draw. The role of the knights tracks will lead us to a constructive suggestion of a new version of chess.

### The eleventh game

```
 1. Ng1-h3    Nb8-c6
 2. Nh3-f4    Ng8-f6
 3. Nf4-d3    d7-d6
 4. Nd3-f4    e7-e5
 5. Nf4-h3    h7-h6
 6. Nh3-g1    Nc6-b4
 7. Nb1-a3    Bc8-e6
 8. Na3-b1    Nb4xa2
 9. Ng1-f3    Be6-d5
10. Nf3-g1    Bf8-e7
11. Ng1-f3    Na2xc1
12. Qd1xc1    o-o
```

13. d2-d3      Bd5xf3
14. e2xf3      Nf6-d5
15. Nb1-c3     Be7-g5
16. Qc1-d1     Nd5xc3
17. b2xc3      Qd8-d7
18. g2-g3      Qd7-c6
19. c3-c4      b7-b5
20. c4xb5      Qc6xb5
21. Bf1-g2     Qb5-b4+
22. Ke1-e2     Bg5-f6
23. Rh1-e1     e5-e4
24. Ra1-b1     e4xd3+
25. Qd1xd3     Ra8-e8+
26. Ke2-f1     Re8xe1+
27. Rb1xe1     Bf6-c3
28. Re1-d1     Rf8-e8
29. Kf1-g1     Re8-e1+
30. Rd1xe1     Bc3xe1
31. f3-f4      Qb4-d2
32. Bg2-e4     Be1xf2+
33. Kg1-g2     Qd2xd3
34. Be4xd3     Bf2-d4
35. Kg2-f3     a7-a5
36. Kf3-e4     Bd4-g1
37. h2-h3      a5-a4
38. Bd3-c4     a4-a3
39. g3-g4      c7-c6
40. Bc4-a2     d6-d5+
41. Ke4-e5     Bg1-e3
42. f4-f5      Be3-c5
43. Ba2-b3     d5-d4
44. Bb3-a2     Kg8-f8
45. Ba2-b3     Kf8-e7
46. Bb3-a2     Bc5-b6
47. Ba2-b3     Bb6-a7
48. Bb3-a2     c6-c5
49. h3-h4      Ba7-b8+
50. Ke5-d5     Bb8-d6
51. g4-g5      h6xg5
52. h4xg5      Ke7-d7
53. g5-g6      f7xg6
54. f5xg6      Bd6-e7
55. Ba2-b3     Be7-f8
56. Bb3-a2     Bf8-d6
57. Ba2-b3     Bd6-e7
58. Bb3-a2     Be7-f8

```
59. Ba2-b3    Kd7-e8
60. Kd5-e6    Bf8-e7
61. Bb3-a2    Be7-d8
62. Ba2-b3    Ke8-f8
63. Ke6-d7    Bd8-e7
64. Bb3-a2    c5-c4
65. Ba2xc4    Be7-g5
66. Bc4-a2    Bg5-f4
67. Ba2-b3    Bf4-h2
68. Bb3-a2    Bh2-g3
69. Ba2-b3    Bg3-f2
70. Bb3-a2    Bf2-e3
71. Ba2-b3    Be3-g1
72. Bb3-a2    Bg1-f2
73. Ba2-b3    Bf2-g3
74. Bb3-a2    Bg3-e1
75. Ba2-b3    Be1-d2
76. Bb3-a2    Bd2-g5
77. Ba2-b3    Bg5-e3.
```
Obviously draw.

I tried to realize the idea of 10 "almost free moves" in some more games, but early attacks of Black often become crucial, thus I finally conclude that 10 such moves as really the *maximum* against this Program.

Probably, for chess on more than 64 squares, and more figures involved, the number of the strange moves might be increased, and, probably, there should be a connection here between these figures/numbers, "10" and "64", of which the first is close to length of the line of the board, i.e. to the square root of the area, if to simplify the things.

Considering that the long initial tracks of White Knights bother Black to confidently develop initiative, and that for a larger board there would be more place for such tracks, one can suggest, say 10×10 board with 4 knights instead of 2 for each side, and two more pawns for each. (Or, at least, 8×10 with the same number of figures as now.) Such a game at the initial stage would look for White like a Corrida Bullfight, if White is obliged to return to initial position. Really an interesting target!

### 2.11. *Some other attempts of the "generous" start, and the "principle of symmetry" for the two-side play of the Program in the confusion state*

I also tried some other "generous" (or half-generous) starts, not based on the "dance" of the white Knights. All of them were less elegant as regards the basic idea, and I would not recommend them for such an experiment.

In one of them, White started with d3 and then Qd1-d2-d1-d2 …. Soon, one of the moves Qd1-d2 was responded to by Black by the unexpected Ng8-h6. The next move of this Knight to the square g4 explained all, -- the sweetness of the square f2 was prevailing, and Black just used the Queen at d2 not letting Bc1xh6. I found this "killing straightforwardness" of Black unattractive.

Another attempt was b2 and g2 and then Bc1-b2-c1 … and Bf1-g2-f1 … . This led to a mostly very difficult (and thus non-recommended) game, and at a certain stage to a very difficult one to evaluate position in which White had two light figures against Rook and two pawns of Black.

Last, but not least, I returned to the idea of the fourth game (Section 2.3) and was trying to let the Program play for both sides, but now *immediately* after the reconstruction, i.e. starting from the move number $N+1$. My impression is that in such positions my patient approach is better for White than the energetic play of the Program for both sides. The Program makes White too active, which is not justified by its poor development, and I observed that White sometimes quickly gets into trouble.

This means that the Program has a "two sided" problem in estimating the strange position, i.e. for the Position of Black confused, the program does not play well for either side. This is not strange, in fact, because the Program thinks also for the opposite side, and it is not so important which side of the board belongs to it.

The academic question of whether we can use the Program for "self criticism" remains open, but we gained the "principle of symmetry" saying that in the state of confusion the Program will play poorly for each side, which is an essential point.

However, let us be complimentary to the Program and show its following "successful" game, where Autoplay was used starting from the seventh move, and White won in a rather combinatory play, not in my style.

This is the "successful" game:

    Twelfth game

1. Ng1-f3    d7-d5
2. Nf3-g1    Ng8-f6
3. Nb1-c3    d5-d4
4. Nc3-b1    Nb8-c6
5. Nb1-a3    e7-e5
6. **Na3-b1**    Nf6-g4
7. f2-f3     Ng4-f6
8. e2-e4 Bf8-e7
9. Bf1-b5 o-o
10. Bb5xc6 b7xc6
11. Ng1-e2  Bc8-e6
12. o-o Ra8-b8
13. d2-d3 c6-c5
14. f3-f4 Qd8-d6
15. f4xe5 Qd6xe5
16. c2-c3 Be7-d6
17. Bc1-f4 Qe5-h5
18. c3xd4 Be6-g4
19. Nb1-c3 Rb8xb2
20. Bf4xd6 c7xd6

21. Qd1-c1 Rb2xe2
22. Nc3xe2 Bg4xe2
23. Rf1-f5 Qh5-g4
24. Rf5-g5 Qg4-h4
25. d4xc5 d6xc5
26. Rg5xc5 Be2xd3
27. Rc5-c8 Qh4xe4
28. Rc8xf8+ Kg8xf8
29. Qc1-a3+ Kf8-e8
30. Qa3xa7 Qe4-e5
31. Ra1-c1 Bd3-f5
32. Qa7-a8+ Ke8-e7
33. Qa8-a7+ Ke7-f8
34. Qa7-a8+ Nf6-e8
35. Rc1-d1 Bf5-g4
36. Rd1-b1 Bg4-d7
37. Kg1-h1 Bd7-f5
38. Rb1-d1 Bf5-c2
39. Rd1-f1 f7-f5
40. Qa8-d8 Qe5-e2
41. Rf1-g1 Qe2-d3
42. Qd8-h4 Ne8-f6
43. Qh4-f2 Bc2-d1
44. Rg1-f1 Bd1-g4
45. Rf1-c1 Nf6-e4
46. Rc1-c8+ Kf8-f7
47. Qf2-a7+ Kf7-g6
48. Rc8-c1 Bg4-d1
49. Rc1-c6+ Ne4-f6
50. Rc6-c7 Qd3-f1+
51. Qa7-g1 Qf1xg1+
52. Kh1xg1 Nf6-d5
53. Rc7-d7 Nd5-e3
54. Kg1-f2 f5-f4
55. g2-g3 Ne3-g4+
56. Kf2-g1 f4xg3
57. h2xg3 Bd1-c2
58. Rd7-d2 Bc2-b1
59. a2-a4 Ng4-e5
60. Rd2-d6+ Kg6-f5
61. a4-a5 Kf5-g4
62. Kg1-f2 Ne5-d3+
63. Kf2-g2 Nd3-c1
64. Rd6-d4+ Kg4-f5
65. a5-a6 Nc1-e2
66. Rd4-b4 Ne2-c3

       67. a6-a7 Bb1-a2
       68. Rb4-d4 Nc3-b5
       69. a7-a8=Q Nb5xd4
       70. Qa8xa2 Kf5-e5
       71. Qa2-g8 Nd4-e6
       72. Qg8xh7 g7-g5
       73. Qh7-d3 Ke5-f6
       74. Kg2-f3 Kf6-e5
       75. Kf3-g4 Ne6-c5
       76. Qd3-f5+ Ke5-d4
       77. Kg4-f3 Kd4-c4
       78. Kf3-e3 Nc5-b3
       79. Qf5xg5 Kc4-b4
       80. g3-g4 Nb3-c5
       81. Qg5-d5 Nc5-a6
       82. Ke3-d4 Resigns

### *2.12. An overview*

The general impressions are as follows:

*The effectiveness of the psychological start is increased by the number of "almost free moves" given to Black. This is natural since the basic idea is to start the development of White using the closeness of Black, and in order to be really close, the black figures need a sufficient number of moves. However, with the increase in the number of "almost free moves" it becomes easier for Black to start an attack and thus to force White to stop being generous. Thus, the tactic of White is to carefully watch the threats of Black and still make it possible to continue to "invite" Black to be closer.*

*For this Program, this tactic cannot continue for more then 10 moves, and not only because there are more possibilities for Black to start an attack. The point is also that after so many moves, the closely approaching Black already succeeds in coordinating its figures, and though the closeness still can be used by White for a quick development, this development may be more difficult and require more skills than in the other cases. White can be strangled.*

The overall impression is illustrated by the following graph (Fig. 4). The size of a dot reflects the probability to at all come to the associated realization of the generous strategy for this number of moves, and the height (in some relative "intuitive" units) of the dot reflects the effectiveness of the defeat, conditionally the generous strategy goes well during these moves.

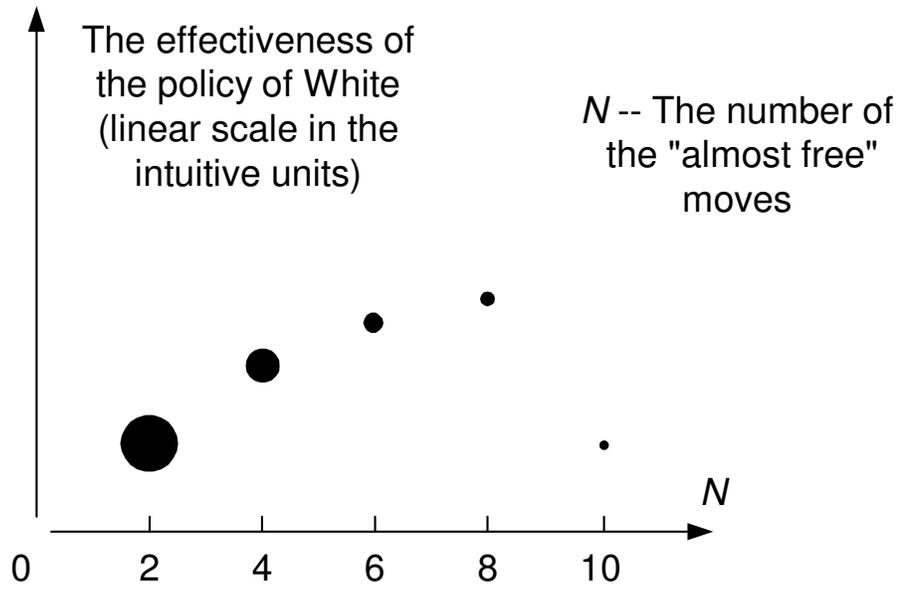

Fig. 4: The intuitive summary of the effectiveness of giving Black the "almost free moves" by starting with the "dance" of the White knights. Up to $N = 8$ there is an increase in the effectiveness, but with a less and less chance of reaching a higher number. At $N = 10$ we have a steep fall of the effectiveness, for the explained reasons.

### *2.13. A modeling of the catastrophic result*

The possibility of missing an early attack of Black, i.e. the attempt to have $N$ too large, can be modeled to some extent, by the mechanical fixture shown in Fig 5, in which we have two parallel, rolling in the opposite directions, rods and a desk lying on them.

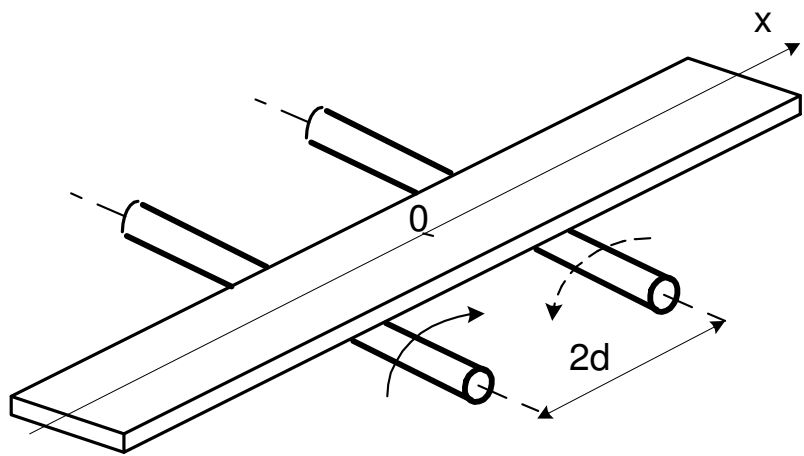

Fig. 5: The massive (of mass $M$) desk on two rods. The left rod (our Black) rotates in the clockwise direction, and right rod (our White) *can* rotate in the counter-clockwise direction, but also can (when it is disconnected from its driver) allow the desk to rotate it in the clockwise

direction. *In the simplest case*, the friction between the rods and the desk can be dry (Coulomb) friction, that is, the tangential friction force is proportional to the normal pressure $N_M$ at the contact, disregarding the value of the relative velocity *if this relative velocity is nonzero. The sign* of this velocity determines the direction of the friction force. Thus, for instance, if the center of the gravity (denoted below as $x_c$) of the desk is closer to the left rod (i.e. $-d < x_c < 0$) then the pressure on this rod is stronger and the friction force applied by it to the desk in the right direction, is stronger than the opposite force, developed in the contact of the desk with the right rode. Qualitatively, such a situation would be also applied to other kinds of physical frictions; the Coulomb friction case is easy to solve and it shows that this is an oscillator.

Consider the simplest case of Coulomb friction. Then, for the friction force we have

$$F(v) = \mu N_M \text{ sign } v_{cr} \qquad (1)$$

where sign $v = 1$ for $v > 0$, and $-1$ for $v < 0$, $\mu$ is the coefficient of the friction, and $v_{cr}$ is *not* just $v = dx_c/dt$, but the *relative velocity* of the desk with respect to the surface of the rotating rod. One sees, that when the rods are indeed rotating as shown in Fig. 5, the friction force (applied to the desk) of the left rode is always directed to the right, and the friction force of the right rod, -- always to the left. That is, though sign $v$ can be positive or negative in the oscillation process studied below, sign $v_{cr}$ will be always positive for the left rod, and always negative for the right rod. Simply, during the oscillatory movement of the desk, each of the forces (rods) sometimes accelerates it, and sometimes brakes. This explains the (constant) signs in the right-hand side of equation (4) below.

For the dry friction, the desk will perform sinusoidal oscillations because the returning summing force is directly proportional to $x_c$ and thus the system is, basically as a mass and a spring. However, -- however nice the sinusoidal oscillations are, the case of dry friction is mainly useful for remembering the system's structure and for seeing its oscillatory nature. We are interested only in *one* pulse of the oscillations oscillation when the desk first moves forward, i.e. $v = dx_c/dt > 0$, and then (hopefully, see below) back, $v < 0$. That Black wins by means of an early attack, means in the model that the desk falls on the right side (when $x_c > d$).

The mathematical theory of this nice system is simple; the nontrivial nuance is just that we obtain an (absolutely precise) linear equation not as a result of an asymptotic smallness of the amplitude of the oscillations, as is usually, but in rigid boundaries for this amplitude. (A system theory specialist could speak about "structural stability", or "robustness" of the linearity.) This "singularity" of the bounds reflects the "catastrophe" result that can occur in the game when White plays too risky (or, takes some certain risk for too long a time).

Below, $P$ denotes the weight of the desk, i.e. $P = Mg$, where $M$ is the mass of the desk, and $N_M^{(1)}$ and $N_M^{(2)}$ are the respective normal pressures caused by the desk on the rods.

Since the desk is not falling and not rotating, we have:

$$N_M^{(1)} + N_M^{(2)} = P, \qquad (2)$$

and

$$N_M^{(1)}(d + x_c) - N_M^{(2)}(d - x_c) = 0. \qquad (3)$$

From these equations,

$$N_M^{(1)} - N_M^{(2)} = -P\frac{x_c}{d},$$

and Newton's equation describing movement of the center of gravity along the *x*-axis,

$$M\frac{d^2 x_c}{dt^2} = \mu N_M^{(1)} - \mu N_M^{(2)}, \qquad (4)$$

becomes

$$M\frac{d^2 x_c}{dt^2} = -\mu P \frac{x_c}{d} = -\mu M g \frac{x_c}{d},$$

or, finally,

$$\frac{d^2 x_c}{dt^2} + \omega_o^2 x_c = 0 \qquad (5)$$

with the cyclic frequency

$$\omega_o = \sqrt{\mu \frac{g}{d}}$$

of the sinusoidal oscillations.

The equation of the usual oscillator is obtained because the saturation of the dry-friction force is equivalent here to a constant gravitational field.

According to (5), the amplitude of the oscillations is constant, obviously, and the *equivalent* (since the desk has only kinetic energy which is not constant) "oscillatory energy" is conserved.

This description is relevant, however, only for $|x_c| \le d$. $|x_c| > d$ means a catastrophe. In chess, the *equality* $|x_c| = d$ (say, black Queen or a Knight, at f2, or a Knight at c2) already means the catastrophe, and for the modeling (with the parameter $\Delta$ introduced below), the permitted boundary value should be less than $d$, $|x_c| < d$.

Risky play of White can be expressed in this model by *delay* in the operation of the right rod. If this rod is disconnected from its driver (our initial "generous" policy), then the desk can rotate it, and will not stop, finally falling at the right side.

Comment: The rotation of the right rod by the desk can be taken into account as an increase in the mass of the desk with the addition which is proportional to the moment of inertia of the rode. This somewhat decreases the frequency of the oscillations, which is not very important here, because our topic is just movement forward and back. This may be of some interest for modeling the chess situation in a students' laboratory. In this laboratory, one can also introduce a switch providing that only at, say, $x_c = d/2$, the right rod automatically starts to be connected to its driver.

   Thus, let the initial conditions be zero, $x_c = 0$, and $v = 0$. The left rod (our Black) starts to act, but the right rod remains passive (this is not, of course, the case of equation (5) that would give zero solution, just delete *for some time* the term $-\mu N_M^{(2)}$ in (4), and thus come to some positive initial conditions for (5)), and is rotated by the desk moving to the right. The right rod will start to rotate in the counter-clock direction only after some delay $\Delta > 0$. The question is what is the upper boundary for $\Delta$ in order not to let $x_c = d$ occur, and start to push the desk back. That is, what is the average optimal number $N$ of "free moves" that White can allow to itself, avoiding the catastrophe?

   That *when the situation does not become catastrophic*, the back movement of the desk will occur at some stage is clear from the fact that for Coulomb friction we can have an oscillatory system.

   However, the chess reality obviously requires some more complicated model of the friction, or (this is much more interesting) a model with cavity in the desk and a massive body (say, a ball or some liquid) with certain freedom of movement in the cavity, which both cannot be develop here. We thus shall be limited by only the above formulation of the statement of the "$\Delta$–problem", i.e. the problem of finding physical models for the analysis of the risky behavior.

   Finishing with the diary, let us continue with its "on-line" observations and with the initial discussion attempting to see in chess not just a competitive game.

## 3. Discussion and conclusions

### 3.1. On the concept of the "best move"

   Though the Reader can assume that the following argument is "put forward" by the very unusual game situation in focus, the point raised is rarely discussed, and it is indeed worth stressing that the concept of "best move" lacks many aspects that are just needed in order to see the game in a wide context.

   In his commentaries on the games of grandmasters [8], Anatoly Karpov says several times: "*The game enters the stage of unobservable complications*", and it seems to be important here also to consider the problematicity of the use of the concept of the "best move", because apart from the rare cases when the Program obviously waits for (anticipates) a typical elementary mistake, it should be seeking the "best move".

   My general old observation (impression) re chess, further supported by the present investigation, is that most chess positions have no "best move". The logical problem is that we can point at the "best move" in an *understood* position, but this understanding will be never complete until we see/find this "best move". Though the concept "best move" is applicable to many positions, this quite objective "faulty logical circle" makes, in general, chess strategy not quite deterministic; the chess position usually is some poorly defined situation, not adjusted to any standard optimization in terms of unique functions. The decision that a move is good (signed as "!" or "!!") is sometimes justified by the final victory, but the decisions are sometimes changed by later analysis.

   Of course, the development of the art of chess is naturally done via well-analyzed positions with best moves found post factum. However, the "number" of the chess positions having the "best move", compared to the positions not having it, seems to be something like the power of a countable set compared to that of a continuum. That is, we can have as much as needed of positions with a best move, helpful for any didactic chess-learning, but these positions are extremely rare among all the possible positions.

### *3.3.  Summary and questions*

1. We have generalized Alyochin's defense to an *Alyochin-type start*, giving in it initiative to the unusually playing White. Based on our experiment, we see such a strategy as a disarming the opponent. It takes the Opponent (the Program) out from the "library", also making him (it) confused for a long time because of having the wrong impression about your real strength, and because of difficulty in returning return to the library sufficiently quickly. Most paradoxically, such a passive defense of White often does not seem to be *objectively* weak, because the undeveloped position of White finally aids (via simplicity of the targets, and the confusion of Black) further development. The sixth game demonstrates that the taking Black out of the library does not necessarily cause unjustified attacks, just a very indecisive play. During the easy development (advance) Black does not take care about good coordination between all of its figures. This is contrary to the case of usual play when such coordination is dictated by the continuous pressure (or resistance) of White.

   Of course, these observations might be incorrect for a stronger program, but the fact is that a programmed *machine can* show clear signs of nervousness, i.e. unjustified early attacks, and depression, i.e. unusually weak play *for many moves* after it is taken out from its library), and the fact is that my scores against the program were strongly improved.

2.  How stable is the use of the (serious) internal library by the program, and how to check this stability most simply? In which cases can we check the stability by asking the program to play, starting from a particular moment, for both sides?

3.  The conclusion that machine is stronger than human player has to be reconsidered, since the psychology can "improve" the human player. Since inclusion of the "psychology" into a program is, in principle, also possible, the conclusions re relative strengths of different programs should be then also reconsidered.

4.  Is the assumption that a Program can be impressed by symmetry of the opponent's constructions correct?

5.   Considering that for a larger board there would be more place for initial confusing tracks of white Knights, we suggest 10×10 board chess game, the "Chess Corrida Bullfight", with 4 knights instead of 2 for each side (or 8×10 with the same figures), in which White is *obliged* to at least once reconstruct its initial position.